\documentclass[runningheads]{llncs}
\usepackage[T1]{fontenc}
\usepackage{graphicx,verbatim}
\usepackage{graphicx}
\usepackage{amssymb}
\usepackage{multirow}
\usepackage{makecell}
\usepackage{amsmath}
\usepackage{colortbl}
\usepackage{xcolor}

\definecolor{lightyellow}{RGB}{135, 206, 235}

\usepackage{hyperref}
\hypersetup{
    colorlinks=false,
    linkbordercolor={1 0 0}, 
    citebordercolor={0 1 0}, 
    urlbordercolor={0 0 1}   
}
\begin{document}
\title{MedKAN: An Advanced Kolmogorov-Arnold Network for Medical Image Classification}
\titlerunning{MedKAN}
%

\author{Zhuoqin Yang\inst{1,2} \and Jiansong Zhang\inst{1} \and Xiaoling Luo\inst{1} \and Zheng Lu\inst{2} \and Linlin Shen\thanks{Corresponding author}\inst{1}}  
\authorrunning{Yang et al.}
\institute{School of Computer Science \& Software Engineering, Shenzhen University, Shenzhen, China \and School of Computer Science, University of Nottingham Ningbo China, Ningbo, China}
    
\maketitle              
\begin{abstract}
Recent advancements in deep learning for image classification predominantly rely on convolutional neural networks (CNNs) or Transformer-based architectures. However, these models face notable challenges in medical imaging, particularly in capturing intricate texture details and contextual features. Kolmogorov-Arnold Networks (KANs) represent a novel class of architectures that enhance nonlinear transformation modeling, offering improved representation of complex features. In this work, we present MedKAN, a medical image classification framework built upon KAN and its convolutional extensions. MedKAN features two core modules: the Local Information KAN (LIK) module for fine-grained feature extraction and the Global Information KAN (GIK) module for global context integration. By combining these modules, MedKAN achieves robust feature modeling and fusion. To address diverse computational needs, we introduce three scalable variants—MedKAN-S, MedKAN-B, and MedKAN-L. Experimental results on nine public medical imaging datasets demonstrate that MedKAN achieves superior performance compared to CNN- and Transformer-based models, highlighting its effectiveness and generalizability in medical image analysis.


\end{abstract}
\section{Introduction}
Medical image classification plays a crucial role in Computer-Aided Diagnosis (CAD) systems, significantly improving diagnostic efficiency and providing automated decision support for clinicians \cite{zhou2021review}. However, this task remains challenging due to the complexity and diversity of medical images \cite{litjens2017survey}. Unlike natural images, pathological features in medical images often exhibit subtle morphological variations, low contrast, and high intra-class similarity, making it difficult to distinguish between different categories. For example, in microscopic images of blood cells, the morphological differences between normal and abnormal cells are often minimal, placing high demands on the feature extraction capabilities of models. In optical coherence tomography (OCT) images of the retina, pathological regions often display irregular shapes and distributions, necessitating global information modeling to differentiate between categories. These challenges impose stringent requirements on deep learning models in terms of feature extraction capability, nonlinear modeling ability, and robustness.

In recent years, deep learning methods have achieved remarkable success in natural image classification tasks and have been widely applied to medical image analysis \cite{chan2020deep}. However, these models have certain limitations when used for medical image classification. On the one hand, CNN-based networks, such as ResNet \cite{he2016deep} and DenseNet \cite{huang2017densely}, perform well in capturing local fine-grained features, such as texture and boundary information of pathological regions. However, their ability to model global features and complex contextual relationships remains limited due to their reliance on local receptive fields. Furthermore, CNNs rely on linear convolution kernels, limiting their ability to model complex nonlinear pathological patterns and reducing their effectiveness in capturing intricate medical image characteristics. On the other hand, Transformer-based networks, such as Vision Transformer (ViT) \cite{dosovitskiy2020vit} and Swin Transformer \cite{liu2021swin}, excel at capturing global information through multi-head self-attention mechanisms. However, their effectiveness often relies on large amounts of labeled data for training and substantial computational resources, which may limit their applicability in medical imaging tasks where annotated data is scarce and real-time processing is required. Recently, the Kolmogorov-Arnold Network (KAN) \cite{liu2024kan,liu2024kan2} has been introduced as a novel deep learning framework, leveraging learnable function approximation and decomposition to enhance nonlinear modeling capabilities. This approach enables KAN to construct efficient representations in high-dimensional feature spaces. However, KAN lacks explicit spatial modeling mechanisms, making it less effective in capturing local pathological structures, which are crucial for medical image classification.

To address these challenges, we propose MedKAN, a deep learning framework specifically designed for medical image classification. By integrating KAN and its convolutional extensions \cite{bodner2024convkan,yang2024activation}, MedKAN effectively captures complex nonlinear pathological structures and enhances feature representation capabilities. Unlike CNNs, which rely on fixed linear convolution operators, MedKAN employs learnable nonlinear transformations to overcome the limitations of CNNs in modeling complex pathological features. Additionally, it reduces dependence on large-scale labeled data, making it more suitable for medical imaging applications with limited computational resources. The key contributions of this work are as follows:
\begin{itemize}
    \item We designed two core modules: the Local Information KAN (LIK) module for fine-grained local feature extraction using grouped KAN-based convolutional networks, and the Global Information KAN (GIK) module for efficient global feature modeling and local-global integration.
    \item We introduced three variants of MedKAN: MedKAN-S, MedKAN-B, and MedKAN-L—by stacking LIK and GIK modules, allowing flexibility in computational resource constraints and task requirements.
    \item We systematically evaluated MedKAN on nine publicly available medical image classification datasets of different modalities. Experimental results demonstrate that MedKAN outperforms state-of-the-art (SOTA) methods in both accuracy and AUC metrics.
\end{itemize}

\begin{figure}[h]
    \centering
    \includegraphics[width=1\textwidth]{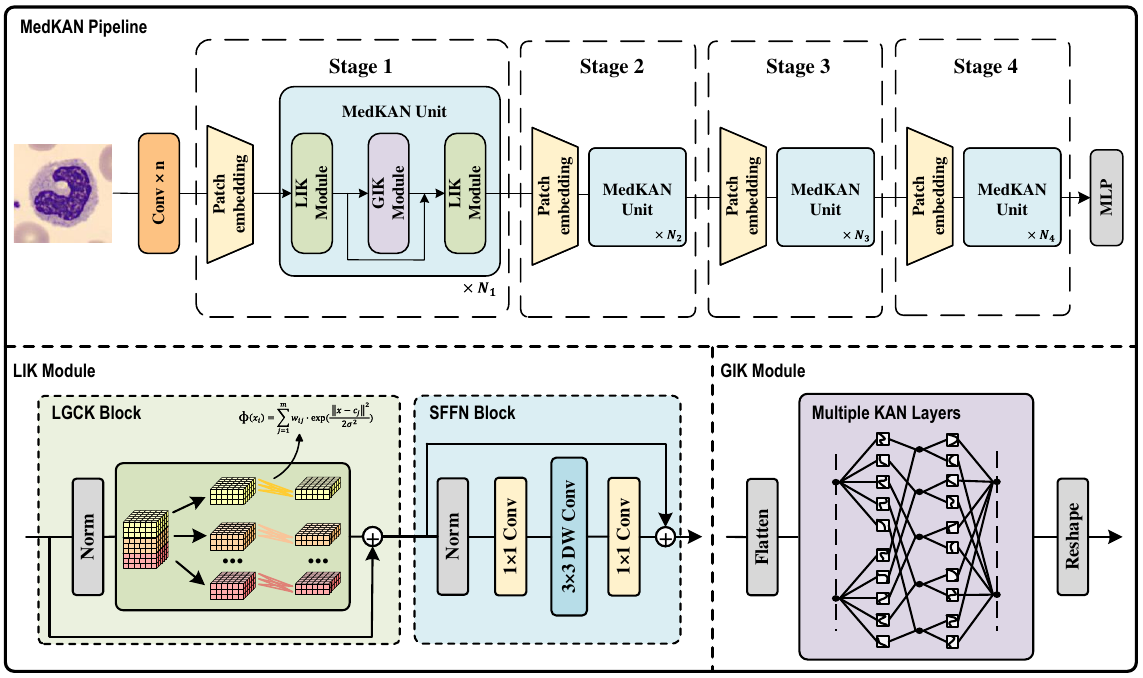}
    \caption{The input medical images are first processed by a basic convolutional module for initial feature extraction and dimensionality reduction, followed by a patch embedding step that further refines the spatial and channel dimensions. The processed features are then processed through a series of stacked modules, including the Local Information KAN (LIK) module for fine-grained feature extraction and the Global Information KAN (GIK) module for capturing long-range dependencies.}
    \label{fig:overview}
\end{figure}

\section{Method}

In this section, we describe the key components of MedKAN, including the Local Information KAN (LIK) module, the Global Information KAN (GIK) module, and the optimizations applied to the KAN layers.





\subsection{LIK Module}

The Local Information KAN (LIK) module enhances MedKAN’s ability to extract fine-grained spatial features. It consists of the Local Grouped Convolution KAN (LGCK) block and the Spatial Feed-Forward Network (SFFN) block, which work together to capture detailed textures and short-range dependencies.

\subsubsection{LGCK Block}

The LGCK block employs grouped KAN-based convolution with residual connections to efficiently model spatial correlations. Unlike traditional convolutions, KAN convolutions replace fixed kernel weights with learnable nonlinear transformation functions, improving the expressiveness of complex patterns \cite{bodner2024convkan,yang2024activation}. Through a grouping mechanism, LGCK maintains computational efficiency in extracting fine-grained features.

Given an input feature map \(\mathbf{F}_{\text{patch}} \in \mathbb{R}^{d \times h \times w}\), the LGCK block divides it into \(g\) groups, each processed by a KAN convolution:

\[
\mathbf{F}_{\text{LGCK}} = \bigoplus_{i=1}^g \text{ConvKAN}(\mathbf{F}_i) + \mathbf{F}_{\text{patch}},
\]
where \(\bigoplus\) denotes the concatenation of grouped features, and \(\text{ConvKAN}_{\text{local}}\) is the grouped KAN convolution, defined as:
\[
\text{ConvKAN}(\mathbf{X}) = \sum_{k=1}^K \phi_k(\mathbf{X}),
\]
with \(\phi_k(\cdot)\) being the learnable nonlinear transformation functions of KAN.

\subsubsection{SFFN Block}

The SFFN block utilizes depthwise separable convolutions (DW Conv) and \(1 \times 1\) convolutions, along with residual connections, to enhance spatial feature learning while maintaining efficiency:

\[
\mathbf{F}_{\text{SFFN}} = \mathbf{F}_{\text{input}} + \text{Conv}_{1 \times 1}(\text{DWConv}_{3 \times 3}(\text{Conv}_{1 \times 1}(\mathbf{F}_{\text{input}}))).
\]

This design improves local feature refinement without significantly increasing computational complexity \cite{sandler2018mobilenetv2}.

\subsection{GIK Module}

The Global Information KAN (GIK) module is designed to model long-range dependencies and global contextual relationships. While the LIK module focuses on fine-grained local textures, GIK ensures that features across distant regions are effectively integrated, which is particularly important in medical imaging where pathological patterns often span large areas.

To achieve efficient global feature modeling, the GIK module replaces traditional multi-head self-attention mechanisms with KAN layers, which reduces computational complexity while preserving strong representational power. The input feature map is first flattened into a sequence of shape \(\mathbf{F}_{\text{flatten}} \in \mathbb{R}^{d \times hw}\), allowing each spatial location to be treated independently while capturing long-range correlations. The sequence is then processed by multiple stacked KAN layers to model nonlinear global interactions:

\[
\mathbf{F}_{\text{GIK}} = \text{KAN}_{\text{global}}(\mathbf{F}_{\text{flatten}}).
\]

By leveraging KAN layers instead of self-attention, GIK effectively integrates global information with a lower computational burden. This design enhances MedKAN’s ability to fuse local and global features, improving its overall representation for medical image classification.

\subsection{KAN Activation Function and Parallel Optimization}

In the original KAN framework, traditional B-spline basis functions were used as activation functions. However, B-spline functions are defined recursively, making them challenging to fully parallelize. This recursive dependency hinders computational efficiency, particularly in high-dimensional medical image processing. 

To address this issue, we adopt the Radial Basis Function (RBF) \cite{buhmann2000radial} as the nonlinear activation function in all KAN layers and convolutional KAN layers. The RBF function is defined as:
\[
\text{RBF}(x) = \exp\left(-\frac{\|x - c\|^2}{2\sigma^2}\right)
\]
where \( c \) is the center and \( \sigma \) is the standard deviation. Unlike B-spline functions, RBF computes each basis function value directly through a mathematical expression, eliminating dependency issues \cite{li2024fastkan}. This characteristic makes the RBF function naturally suitable for GPU-based parallelization. This optimization significantly improves the processing speed of MedKAN when handling high-dimensional medical image data.

\section{Experiments \& Results}

In this section, we describe the datasets, implementation details, evaluation metrics, experimental results comparing our proposed model with existing baseline and state-of-the-art (SOTA) methods, and the ablation study.

\subsection{Datasets and Evaluation Metrics}

To evaluate the generalizability of MedKAN on diverse medical datasets, we selected nine classification datasets from MedMNIST \cite{yang2021medmnist,yang2023medmnistv2}. These include 17,092 microscopic blood cell images from BloodMNIST (BlM), 780 chest ultrasound images from BreastMNIST (BrM), 10,015 dermatoscopic images of skin lesions from DermaMNIST (DeM), 109,309 optical coherence tomography images of retinal diseases from OCTMNIST (OCTM), 5,856 pediatric chest X-ray images from PneumoniaMNIST (PnM), and 236,386 kidney cortex cell images from TissueMNIST (TiM). Additionally, we include three tumor CT datasets—OrganAMNIST, OrganCMNIST, and OrganSMNIST (OrAM, OrCM, OrSM)—corresponding to axial (58,830 images), coronal (23,583 images), and sagittal (25,211 images) views, respectively.

These datasets cover classification tasks ranging from binary to 11-class problems. All datasets underwent standardized preprocessing and resizing to 224×224 pixels. The datasets were pre-split into training, validation, and test sets by MedMNIST. Consistent with prior works \cite{yang2023medmnistv2}, we used Accuracy (ACC) and Area under the ROC Curve (AUC) as evaluation metrics.

\begin{table}[h]
\caption{Performance comparison of MedKAN and state-of-the-art methods on diverse medical datasets. A dash (—) indicates that the corresponding value is not provided. \textbf{Bold} indicates the best, \underline{underline} indicates the second-best.}

\label{tab:big}
\centering
\resizebox{\textwidth}{!}{ 
\begin{tabular}{|l|c|c|l|c|c|l|c|c|l|c|c|}
\hline
\multicolumn{3}{|c|}{\textbf{BlM}} & \multicolumn{3}{c|}{\textbf{BrM}} & \multicolumn{3}{c|}{\textbf{DeM}} \\ \hline
\multicolumn{1}{|c|}{Method} & ACC & AUC & \multicolumn{1}{|c|}{Method} & ACC & AUC & \multicolumn{1}{|c|}{Method} & ACC & AUC \\ \hline
AutoML\cite{bisong2019googleauto}\scriptsize{Apress'19}& 0.966 & \underline{0.998} & SADAE\cite{ge2022sadae}\scriptsize{TCSVT'22} & 0.859 & 0.897 & MedViT-S\cite{manzari2023medvit}\scriptsize{CIBM'23} & 0.780 & 0.937 \\ 
MedViT-L\cite{manzari2023medvit}\scriptsize{CIBM'23} & 0.954 & 0.996 & BP-CapsNet\cite{lei2023bp}\scriptsize{Appl.S.C'23} & 0.840 & 0.824 & BP-CapsNet\cite{lei2023bp}\scriptsize{Appl.S.C'23} & 0.774 & 0.923 \\ 
BP-CapsNet\cite{lei2023bp}\scriptsize{Appl.S.C'23} & 0.946 & 0.996 & MedViT-S\cite{manzari2023medvit}\scriptsize{CIBM'23} & \underline{0.897} & \underline{0.938} & Med-Former\cite{chowdary2024medformer}\scriptsize{MICCAI'24} & \underline{0.783} & \underline{0.946}\\ 
Med-Former\cite{chowdary2024medformer}\scriptsize{MICCAI'24}   & 0.965 & 0.997& EHDFL\cite{han2023ehdfl}\scriptsize{CIBM'23} & \underline{0.897} & 0.894 & HideMIA\cite{lin2024hidemia}\scriptsize{ACM MM'24} & 0.781 & 0.924 \\
\rowcolor{lightyellow}
MedKAN-S & 0.968 & \underline{0.998} & MedKAN-S  &\textbf{0.904} & \textbf{0.942} & MedKAN-S  & 0.776 & 0.926 \\ 
\rowcolor{lightyellow}
MedKAN-B & \textbf{0.986} & \textbf{0.999} & MedKAN-B  & 0.895 & 0.937 & MedKAN-B& \textbf{0.803} & \textbf{0.950}\\
\rowcolor{lightyellow}
MedKAN-L & \underline{0.979} & \textbf{0.999} & MedKAN-L  & 0.885 & 0.859& MedKAN-L& 0.780 & 0.938\\ \hline
\multicolumn{3}{|c|}{\textbf{OrAM}} & \multicolumn{3}{c|}{\textbf{OrCM}} & \multicolumn{3}{c|}{\textbf{OrSM}} \\ \hline
\multicolumn{1}{|c|}{Method} & ACC & AUC & \multicolumn{1}{|c|}{Method} & ACC & AUC & \multicolumn{1}{|c|}{Method} & ACC & AUC \\ \hline
ResNet-18\cite{he2016deep}\scriptsize{CVPR'16} & \underline{0.951} & \textbf{0.998} & ResNet-18\cite{he2016deep}\scriptsize{CVPR'16} & 0.920 & \underline{0.994} & ResNet-50\cite{he2016deep}\scriptsize{CVPR'16} & 0.785 & 0.975 \\ 
MedViT-L\cite{manzari2023medvit}\scriptsize{CIBM'23}   & 0.943 & \underline{0.997} & MedViT-L\cite{manzari2023medvit}\scriptsize{CIBM'23} & 0.922 & 0.994 & AutoKeras\cite{jin2019auto}\scriptsize{SIGKDD'19} & \underline{0.813} & 0.974 \\ 
EHDFL\cite{han2023ehdfl}\scriptsize{CIBM'23}  & 0.936 & \underline{0.997} & EHDFL\cite{han2023ehdfl}\scriptsize{CIBM'23} & 0.916 & 0.994 & MedViT-L\cite{manzari2023medvit}\scriptsize{CIBM'23} & 0.806 & 0.973\\ 
IncARMAG\cite{remigio2025incarmag}\scriptsize{NCompu.'25}  & 0.927 & 0.993 & IncARMAG\cite{remigio2025incarmag}\scriptsize{NCompu.'25} & \underline{0.934} & \textbf{0.995} & IncARMAG\cite{remigio2025incarmag}\scriptsize{NCompu.'25} & 0.801 & 0.975\\ 
\rowcolor{lightyellow}
MedKAN-S & 0.950 & \underline{0.997} & MedKAN-S & 0.912 & 0.993 & MedKAN-S & 0.799 & 0.973\\ 
\rowcolor{lightyellow}
MedKAN-B & \textbf{0.962} & \textbf{0.998} & MedKAN-B & \textbf{0.940} & \textbf{0.995} & MedKAN-B & \textbf{0.815} & \textbf{0.979} \\ 
\rowcolor{lightyellow}
MedKAN-L & 0.938  & 0.996 & MedKAN-L & 0.926 & \underline{0.994} & MedKAN-L & 0.808  & \underline{0.977}\\ \hline
\multicolumn{3}{|c|}{\textbf{OCTM}} & \multicolumn{3}{c|}{\textbf{PnM}} & \multicolumn{3}{c|}{\textbf{TiM}} \\ \hline
\multicolumn{1}{|c|}{Method} & ACC & AUC & \multicolumn{1}{|c|}{Method} & ACC & AUC & \multicolumn{1}{|c|}{Method} & ACC & AUC \\ \hline
ResNet-50\cite{he2016deep}\scriptsize{CVPR'16} & 0.776 & 0.958 & AutoML\cite{bisong2019googleauto}\scriptsize{Apress'19} & 0.946 & 0.991 & AutoKeras\cite{jin2019auto}\scriptsize{SIGKDD'19} & 0.703 & 0.941 \\ 
MedViT-L\cite{manzari2023medvit}\scriptsize{CIBM'23}   & 0.791 & 0.945 & SADAE\cite{ge2022sadae}\scriptsize{TCSVT'22} & 0.901 & 0.983 & Twin-SVT\cite{chu2021twins}\scriptsize{NeurIPS'21} & 0.721 & — \\ 
EHDFL\cite{han2023ehdfl}\scriptsize{CIBM'23}  & 0.824 & 0.972&  BP-CapsNet\cite{lei2023bp}\scriptsize{Appl.S.C'23} & 0.920 & 0.970 & ConvNext\cite{liu2022convnet}\scriptsize{CVPR'22} & 0.726 & — \\ 
IncARMAG\cite{remigio2025incarmag}\scriptsize{NCompu.'25}   & 0.873 & 0.985 & MedViT-S\cite{manzari2023medvit}\scriptsize{CIBM'23} & \underline{0.961} & \underline{0.995} & MedViT-S\cite{manzari2023medvit}\scriptsize{CIBM'23} & \underline{0.731} & \underline{0.952}\\ 
\rowcolor{lightyellow}
MedKAN-S & 0.921 &0.993 & MedKAN-S & 0.952 & 0.993 & MedKAN-S & 0.720 & 0.946 \\
\rowcolor{lightyellow}
MedKAN-B & \textbf{0.927} & \textbf{0.996} & MedKAN-B & \textbf{0.963} & \textbf{0.996} & MedKAN-B & \textbf{0.740} & \textbf{0.957}\\ 
\rowcolor{lightyellow}
MedKAN-L & \underline{0.925} & \underline{0.994} & MedKAN-L & 0.924 & 0.988 & MedKAN-L & 0.726 & 0.949 \\ \hline
\end{tabular}
}
\end{table}

\subsection{Implementation Details}

Our experiments were conducted on the nine diverse datasets mentioned above. We followed settings similar to MedMNISTv2 \cite{yang2023medmnistv2} to ensure fair comparisons. Specifically, cross-entropy loss was employed during training with a batch size of 64. The Adam optimizer \cite{kingma2014adam} was used with a learning rate of 0.0001 and a weight decay of 0.0001. Each model was trained for up to 150 epochs, with early stopping applied to prevent overfitting. Multiple runs were conducted, and the average results were reported. All experiments were performed on an NVIDIA RTX A6000 GPU with 48GB of RAM. Furthermore, we introduced three variants of the model, MedKAN-S, MedKAN-B, and MedKAN-L, by adjusting the stacking of LIK and GIK blocks and the embedding dimensions.

\subsection{Experimental Results}

\begin{figure}[h]
    \centering
    \includegraphics[width=1\textwidth]{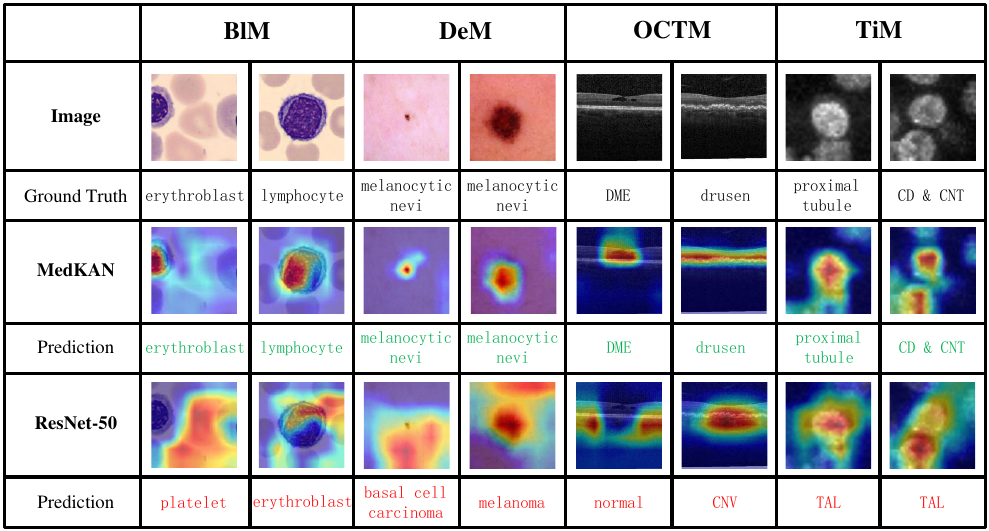}
    \caption{Grad-CAM visualizations and classification predictions of MedKAN and ResNet-50 on four datasets (BlM, DeM, OCTM, TiM). The ground truth and corresponding model predictions are displayed for each dataset. Grad-CAM heatmaps highlight the regions of interest identified by each model, demonstrating MedKAN’s improved localization and classification accuracy compared to ResNet-50.}
    \label{fig:gradcam}
\end{figure}

We evaluated the proposed MedKAN against other SOTA methods on various MedMNIST datasets using ACC and AUC. As shown in Table \ref{tab:big}, MedKAN outperformed existing methods across all datasets, with MedKAN-B achieving the best performance. Particularly on BloodMNIST and OCTMNIST, MedKAN surpassed the SOTA methods with ACCs of 0.986 and 0.927, showing improvements of 0.020 and 0.054, respectively. These results demonstrate the generalizability and precision of MedKAN across different medical imaging tasks.

\begin{table}[h]
\caption{Comparison of model parameter sizes and accuracy on the TissueMNIST dataset, highlighting MedKAN’s superior performance across different model sizes.}
\label{tab:size}
\resizebox{\textwidth}{!}{ 
\centering
\begin{tabular}{|l|c|c|l|c|c|l|c|c|l|c|c|}
\hline
\multicolumn{3}{|c|}{\textbf{Small Size}} & \multicolumn{3}{c|}{\textbf{Medium Size}} & \multicolumn{3}{c|}{\textbf{Large Size}} \\ \hline
\multicolumn{1}{|c|}{Method} & Para. & ACC & \multicolumn{1}{|c|}{Method} & Para. & ACC & \multicolumn{1}{|c|}{Method} & Para. & ACC \\ \hline
ResNet-18\cite{he2016deep}\scriptsize{CVPR'16} & 11.7M & 0.681 & ResNet-50\cite{he2016deep}\scriptsize{CVPR'16} & 25.6M & 0.680 & ResNet-152\cite{he2016deep}\scriptsize{CVPR'16} & 60.2M & 0.675 \\ 
EfficientNet\cite{tan2019efficientnet}\scriptsize{PMLR'19}  & 12M & 0.690 & CvT-13\cite{wu2021cvt}\scriptsize{ICCV'21} & 20.1M & 0.716 & ConvNeXt\cite{liu2022convnet}\scriptsize{CVPR'22} & 88M & 0.691 \\ 
RVT-Ti\cite{mao2022rvt}\scriptsize{CVPR'22}  & 8.6M & 0.696 & Twins-SVT\cite{chu2021twins}\scriptsize{NeruIPS'21} & 24M & 0.721 & RVT-B\cite{mao2022rvt}\scriptsize{CVPR'22} & 86.2M & 0.693\\ 
MedViT-T\cite{manzari2023medvit}\scriptsize{CIBM'23}   & 10.8M & 0.703 & MedViT-S\cite{manzari2023medvit}\scriptsize{CIBM'23} & 23.6M & 0.731 & MedViT-L\cite{manzari2023medvit}\scriptsize{CIBM'23} & 45.8M & 0.699 \\ 
\rowcolor{lightyellow}
MedKAN-S & 11.5M & \textbf{0.720} & MedKAN-B   & 24.6M & \textbf{0.740} & MedKAN-L  & 48.0M & \textbf{0.726} \\  \hline
\end{tabular}
}
\end{table}

We further compared MedKAN with other parameter-similar models on TissueMNIST, as shown in Table \ref{tab:size}. MedKAN-S, MedKAN-B, and MedKAN-L consistently outperformed their parameter-equivalent counterparts. MedKAN-S offered a lightweight solution while maintaining acceptable performance, particularly on simpler tasks like BreastMNIST. In contrast, MedKAN-L showed slight performance degradation due to its complexity, likely requiring larger datasets. Nonetheless, MedKAN-L still outperformed other models with similar parameter sizes, further validating the applicability of the proposed framework in the medical domain.

\subsection{Ablation Study}

\begin{table}[h]
\caption{Ablation study on different configurations across TiM and BlM datasets.}
\label{tab:ablation}
\centering
\begin{tabular}{|c|c|c|c|c|c|c|c|c|c|}
\hline
\multirow{3}{*}{\makecell{Residual \\ Block}} & \multirow{3}{*}{\makecell{ConvNeXt \\ Block}} & \multirow{3}{*}{\makecell{LIK w/ \\ Conv}} & \multirow{3}{*}{\makecell{LIK w/ \\ KANConv}} & \multirow{3}{*}{\makecell{GIK w/ \\ MLP}} & \multirow{3}{*}{\makecell{GIK w/ \\ KAN}} & \multicolumn{4}{c|}{Datasets} \\ \cline{7-10} 
 &  &  &  &  &  & \multicolumn{2}{c|}{TiM} & \multicolumn{2}{c|}{BlM} \\ \cline{7-10} 
 &  &  &  &  &  & ACC & AUC & ACC & AUC \\ \hline
 \checkmark &  &  &  &  & \checkmark & 0.703 & 0.939 & 0.953 & 0.998 \\ 
 & \checkmark &  &  &  & \checkmark & 0.708 & 0.941 & 0.958 & 0.998 \\ 
 &  & \checkmark &  &  & \checkmark & 0.725 & 0.946 & 0.964 & 0.999 \\ \hline
 &  &  & \checkmark &  &  & 0.723 & 0.946 & 0.969 & 0.999 \\ 
 &  &  & \checkmark & \checkmark &  & 0.728 & 0.947 & 0.976 & 0.999 \\ 
 &  &  & \checkmark &  & \checkmark & \textbf{0.740} & \textbf{0.957} & \textbf{0.986} & \textbf{0.999} \\ \hline
\end{tabular}
\end{table}

We conducted ablation experiments to evaluate the contributions of the LIK and GIK modules using two datasets from MedMNIST (Table \ref{tab:ablation}). Replacing LIK with traditional modules (e.g., Residual Block, ConvNeXt Block) or standard convolution validated the effectiveness of KANConv. Similarly, removing GIK or replacing KAN with MLP demonstrated the importance of global information integration. Results showed that LIK and GIK significantly improved classification performance, with KANConv and nonlinear transformations excelling in feature modeling and generalization. The combined use of LIK and GIK achieved optimal performance on TiM and BlM datasets, validating the effectiveness of MedKAN's architecture.

\section{Conclusion}
In this paper, we proposed MedKAN, a general-purpose medical image classification framework based on the Kolmogorov-Arnold Network (KAN). MedKAN leverages the Local Information KAN (LIK) module and Global Information KAN (GIK) module to effectively capture both local and global image features. By stacking these modules, we developed three variants—MedKAN-S, MedKAN-B, and MedKAN-L—with varying parameter sizes.

Experimental results on nine diverse medical imaging datasets demonstrated that MedKAN consistently outperformed methods based on CNN and Transformer architectures, showcasing its robustness and adaptability across tasks. These findings highlight the potential of KAN in medical image analysis. In the future, we plan to further explore its applications and extend its capabilities in the medical imaging domain.

\bibliographystyle{splncs04}
\bibliography{ref1}

\end{document}